\pdfoutput=1

\documentclass[11pt]{article}

\usepackage[dvipsnames,table,xcdraw]{xcolor}
\usepackage{authblk}
\usepackage[]{acl}

\usepackage{times}
\usepackage{latexsym}

\usepackage[T1]{fontenc}

\usepackage[utf8]{inputenc}

\usepackage{microtype}
\usepackage{graphicx}
\usepackage{subfig}
\usepackage{booktabs} 
\usepackage{mdframed}

\newcommand{\tot}{Tree-of-Traversals }

\usepackage{graphicx}
\usepackage[most]{tcolorbox}
\tcbuselibrary{listings}
\usepackage{caption}

\usepackage{float}

\usepackage{algorithm}
\usepackage[noend]{algpseudocode}
\usepackage{booktabs}
\usepackage{subcaption}

\title{Tree-of-Traversals: A Zero-Shot Reasoning Algorithm for Augmenting Black-box Language Models with Knowledge Graphs}

\author[1,2]{Elan Markowitz\thanks{\ \ Work done while interning at Amazon\\Corresponding author: \texttt{esmarkow@usc.edu}}\ \ $^,$}
\author[1]{Anil Ramakrishna}
\author[1]{Jwala Dhamala}
\author[1]{Ninareh Mehrabi}
\author[1]{\authorcr Charith Peris}
\author[1]{Rahul Gupta}
\author[1]{Kai-Wei Chang}
\author[1]{Aram Galstyan}

\affil[1]{Amazon AGI}
\affil[2]{University of Southern California}

\begin{document}
\maketitle
\begin{abstract}
Knowledge graphs (KGs) complement Large Language Models (LLMs) by providing reliable, structured, domain-specific, and up-to-date external knowledge. However, KGs and LLMs are often developed separately and must be integrated after training. 
We introduce Tree-of-Traversals, a novel zero-shot reasoning algorithm that enables augmentation of black-box LLMs with one or more KGs. The algorithm equips a LLM with actions for interfacing a KG and enables the LLM to perform tree search over possible thoughts and actions to find high confidence reasoning paths. We evaluate on two popular benchmark datasets. Our results show that \tot significantly improves performance on question answering and KG question answering tasks. Code is available at \url{https://github.com/amazon-science/tree-of-traversals}

\end{abstract}

\newcommand{\textcircledcentered}[1]{\raisebox{.5pt}{\textcircled{\raisebox{-.9pt} {#1}}}}

\begin{figure*}[!thb]
\begin{mdframed}[innertopmargin=15, innerbottommargin=0]
\begin{minipage}[t]{.55\textwidth}
\vspace*{-\dimexpr\baselineskip+\heavyrulewidth+\abovetopsep\relax}
\begin{tabular}[t]{m{.3\textwidth} m{.65\textwidth}}
    \subfloat{\includegraphics[width=.3\textwidth]{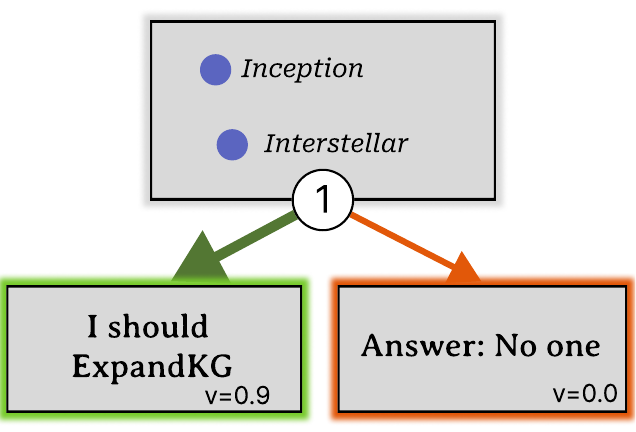}} & \small \textcircledcentered{1} Initial state seeded with the KG entities from the query. Each subsequent step consists of selecting a tree node (rectangle), sampling thoughts or actions from LLM, performing actions, and evaluating outputs with LLM. \\
    \subfloat{\includegraphics[width=.3\textwidth]{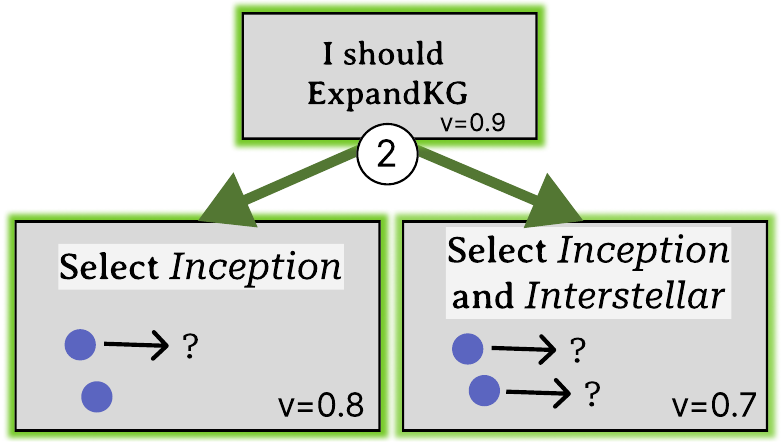}} & \small \textcircledcentered{2} \tot selects which node to search based on the values of the potential options. The ``Expand KG'' option had higher value than ``Answer: No one''. After choosing to Expand KG, the model next selects which entities from the KG to expand. \\
    \subfloat{\includegraphics[width=.3\textwidth]{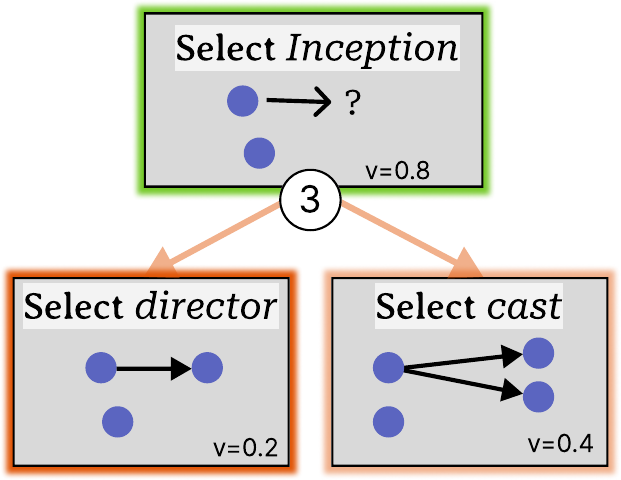}} & \small \textcircledcentered{3} After selecting entities, the model next chooses what relation to expand upon. The local knowledge graph is then updated with the requested information. Unfortunately, in this case, the sampled actions yielded low value output states. \\
\end{tabular}
\end{minipage}
\hfill
\begin{minipage}[t]{.4\textwidth}
\centering
\subfloat{
    \includegraphics[width=\textwidth]{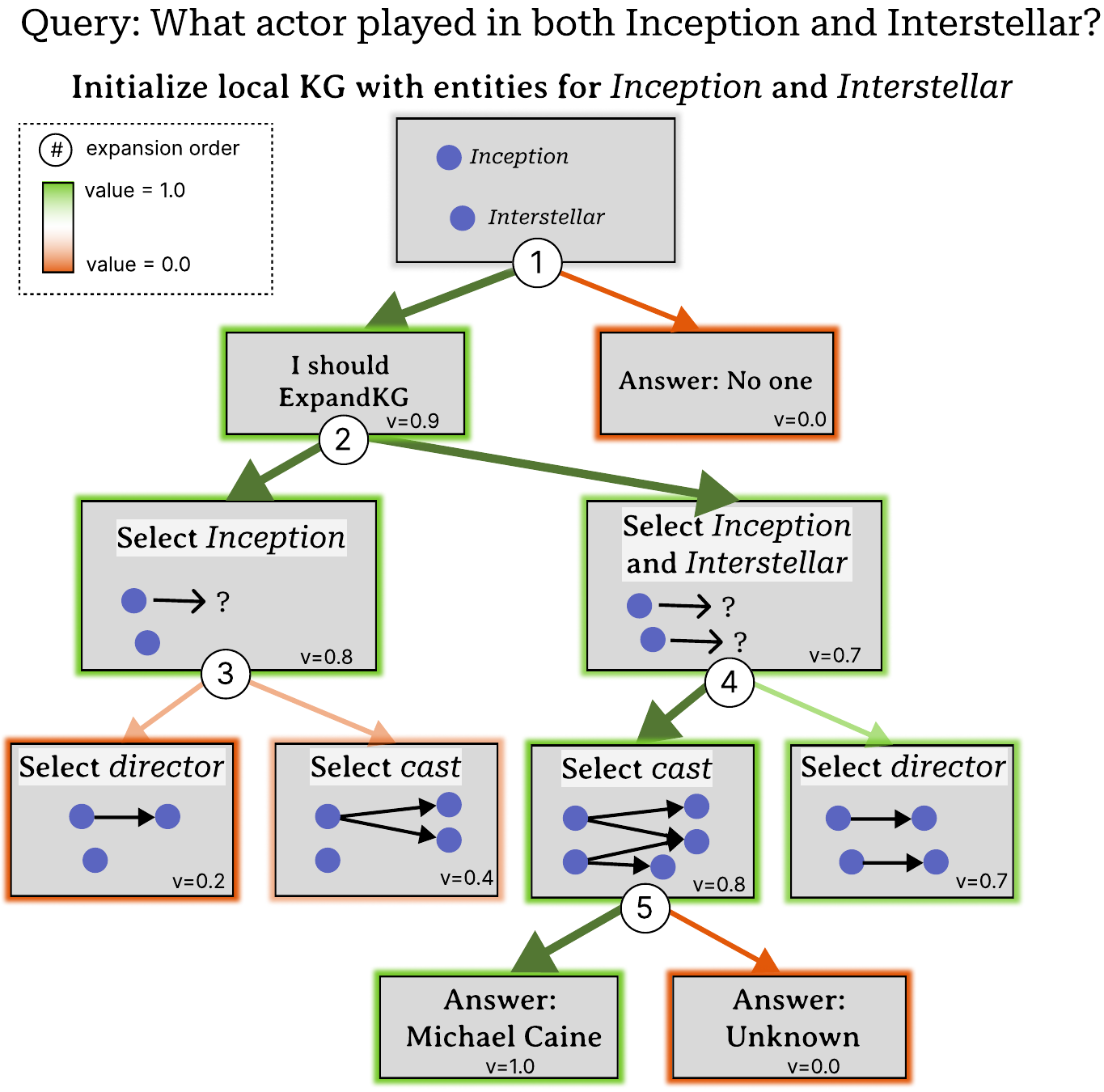}
}
\end{minipage}
\\
\begin{minipage}[t]{\textwidth}
\begin{tabular}{m{.15\textwidth} m{.25\textwidth} m{.15\textwidth} m{.35\textwidth}}
    \subfloat{\includegraphics[width=.15\textwidth]{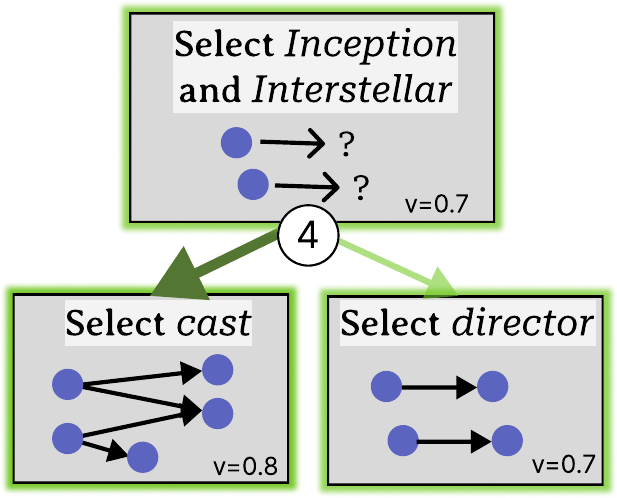}} & 
    \small \textcircledcentered{4} Since the previous selection yielded low-value states, the model backtracks to the best alternative state. & 
    \subfloat{\includegraphics[width=.15\textwidth]{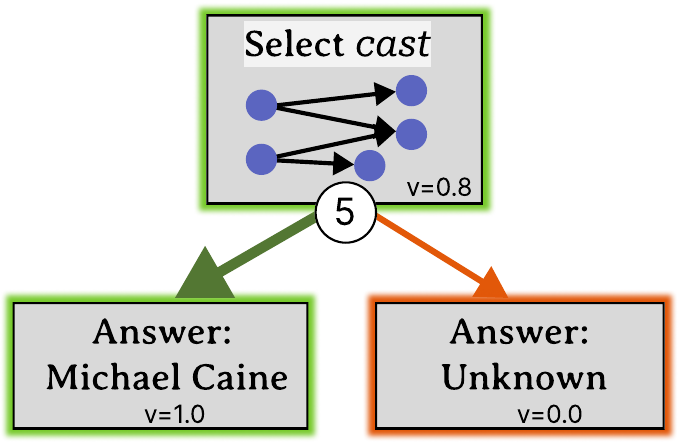}} &
    \small \textcircledcentered{5} \tot now has all the information needed for the query and generates two answers. The correct answer is given a high value by the LLM and is returned to the user. 
\end{tabular}
\end{minipage}%
\end{mdframed}
\caption{An example of how \tot uses a KG interface for the query, ``What actor played in both Inception and Interstellar?''.}
\label{fig:main}
\end{figure*}

\section{Introduction}
Large language models (LLMs) are used for a range of knowledge-intensive tasks such as information retrieval \cite{Zhu2023LargeLM}, summarization \cite{Zhang2023ExtractiveSV}, and question answering \cite{Tan2023CanCR}. Trained on large amounts of textual data, these models learn a wide breadth of information. However, LLMs suffer from several limitations -- they produce hallucinated information~\cite{Ji2022SurveyOH, Bang2023AMM}, lack deep domain-specific knowledge \cite{Pan2023UnifyingLL}, and have a static knowledge cutoff when training ends.

Knowledge graphs (KGs) naturally complement LLM weaknesses. KGs contain up-to-date information covering general~\cite{Vrandei2014WikidataAF, Lehmann2015DBpediaA} and/or domain-specific topics~\cite{AbuSalih2020DomainspecificKG, Liu2019AnticipatingSM, Zhu2017IntelligentLF, Choi2019InferenceOB, Farazi2020KnowledgeGA} in highly structured and interpretable format. Augmenting an LLM's ability to reason and respond in natural language with an external KG's up-to-date knowledge presents a path toward a  reliable and factual LLM. 

The rise of powerful LLMs with new capabilities has renewed interest in combining LLMs with KGS. Numerous survey and position papers have recently emphasized their combined potential \cite{Pan2023LargeLM, Zhu2023LLMsFK, Yang2023ChatGPTIN, Pan2023UnifyingLL}. 
Existing works augmented LLMs with KGs in multiple ways, such as integration into pretraining~\cite{Yasunaga2022DeepBL}, fine-tuning~\cite{Zhang2022GreaseLMGR}, or later adaptation with subsequently trained components~\cite{Lin2019KagNetKG, Hu2022EmpoweringLM}. All of these carry some limitations. In particular, training or fine-tuning a large scale LLMs is computationally expensive. 
In some cases, model weights are unavailable publicly.
Finally, the largest KGs require their own servers and cannot be integrated in-memory with LLMs. Additionally, previous works do not consider the case of augmenting with multiple KGs. 

An algorithm that allows the augmentation of a powerful black-box LLM with any number of internal or external KGs without training from scratch or fine-tuning the model is valuable. Such a zero-shot algorithm would enable several innovative use cases such as (1) customers using a black-box LLM API in conjunction with an internal domain-specific KG, (2) integration of personalized KGs into an LLM without the risks associated with training a model with such personal data, (3) integration of deep domain knowledge via an array of API accessible KGs (e.g., IMDb\footnote{\href{https://aws.amazon.com/marketplace/seller-profile?id=0af153a3-339f-48c2-8b42-3b9fa26d3367}{IMDb API}}, MusicBrainz\footnote{\href{https://musicbrainz.org/doc/MusicBrainz_API}{MusicBrainz API}}).

We introduce \textbf{Tree-of-Traversals}, a novel algorithm that addresses the above issues by allowing the augmentation of any powerful LLM with arbitrary number of KGs in a zero-shot fashion. It requires no training, functions with black-box access to the LLM, and works with any API accessible KG. Our contributions are:
\begin{enumerate}
    \item Tree of traversals: A novel zero-shot algorithm for augmenting any powerful LLM with arbitrary number of KGs and enabling advanced KG reasoning using tree search. 
    \item Evaluation of \tot on two question answering tasks: 2WikiMultiHop and QALD-10 and comparison with baselines.
    \item Development of a new dataset to test combined reasoning over a general and a domain-specific KG, and evaluation of \tot on this dataset.
\end{enumerate}

We conduct detailed experiments on three models of varied sizes hosted on Amazon Bedrock and present detailed ablation studies.

\section{Tree-of-Traversals}
The \tot algorithm maintains a local KG subgraph that is repeatedly expanded until it contains all the information required by an LLM to answer the given query. At the start, a local KG subgraph is initialized to contain the entities present in the original query. It is then expanded using a tree search algorithm to choose actions and thoughts generated by an LLM in order to obtain relevant knowledge from the KG using a KG interface. The algorithm halts when the LLM is able to answer the original query using the local KG subgraph as context. \tot consists of three major components. 
(1) \textbf{A knowledge graph interface} implemented to interact with one or more required KGs.
(2) \textbf{An action state machine (ASM)} which is a finite state machine that defines the feasible space of actions, states, and prompt templates when the LLM interacts with a KG to expand the local KG subgraph.
(3) \textbf{A tree search algorithm} which defines the overall LLM search trajectory such as best first search, backtrack upon making a mistake, and termination condition when an answer is found. 

\subsection{Knowledge Graph Interface}
\label{ssec:kginterface}
The knowledge graph interface allows \tot to interact with one or multiple KGs. Let $\mathcal{K}=(\mathcal{E}, \mathcal{R}, \mathcal{T})$ be a single KG. $\mathcal{E}$ is the set of entities in which each entity consists of an identifier, a label, and an optional description (e.g., Q35332, `\textit{Christopher Nolan}', `\textit{British-American filmmaker}'). $\mathcal{R}$ is the set of relation types, each consisting of an identifier, a label, and an optional inverse label (P57, `\textit{director}', `\textit{is director of}'). $\mathcal{T}$ is the set of edges or facts in the KG in which each edge is of the form $(s, r, o)$ where $s, o \in \mathcal{E}$ and $r\in \mathcal{R}$, e.g., (`\textit{Inception}', `\textit{director}', `\textit{Christopher Nolan}'). \tot can be used for $\mathcal{K}$ as long as the following interfaces are implemented. 
\begin{enumerate}
    \item $\texttt{initialize}(q)\rightarrow \mathcal{E}_0$: It takes as input a query $q$, extracts entities from $q$ and returns the linked entities $\mathcal{E}_0 \subset \mathcal{E}$ where $\mathcal{E}_0$ are the entities from $\mathcal{K}$ that are referenced in $q$. 
    \item $\texttt{get\_relations}(\mathcal{E}_{selected})\rightarrow \mathcal{R}_{options}$:
It takes a set of entities, $\mathcal{E}_{selected}$, and returns the relation types, $\mathcal{R}_{options} \subset \mathcal{R}$, that $\mathcal{E}_{selected}$ have in $\mathcal{K}$: $\left\{r|(s,r,o)\in\mathcal{T},s\in\mathcal{E}_{selected}\right\}$
    \item $\texttt{get\_edges}(\mathcal{E}_{selected}, r)$
$\rightarrow$ 
$\mathcal{T}_{added}, \mathcal{E}_{added}$: It takes a set of entities and a selected relation type, and returns all edges with relation type $r$ for source entities in $\mathcal{E}_{selected}$:
$\left\{(s,r,o)\in \mathcal{T}|s\in\mathcal{E}_{selected}, r=r, o\in \mathcal{E}\right\}$. It also returns the new entities, $\mathcal{E}_{added}$, that are entities reached with $\mathcal{T}_{added}$.
\end{enumerate}

This interface is implemented with SPARQL queries when available; otherwise, we use the graph API that is available for the KG. For multiple KGs, each interface is implemented separately.

\subsection{Action State Machine (ASM)}
\label{ssec:asm}

One of the challenges with developing a zero-shot LLM algorithm that works with arbitrary KGs is that the LLM does not know what relations are available in the graph or what relations are valid for a given entity. Few-shot or in-context learning approaches can only cover a handful of the possible relation types (e.g., Wikidata has over 11,000 relation types)~\cite{Brown2020LanguageMA}. 

\begin{figure}
    \centering
    \includegraphics[width=0.6\columnwidth]{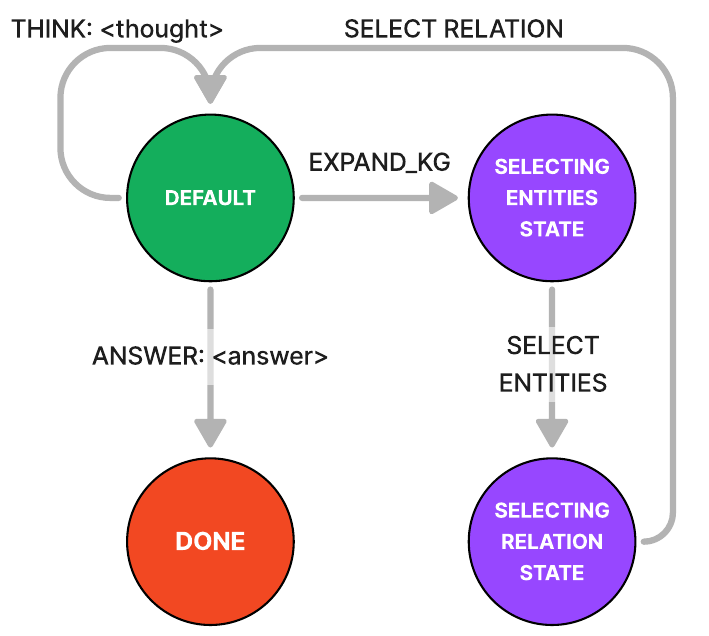}
    \caption{Action State Machine}
    \label{fig:action_state_machine}
\end{figure}

To overcome these issues we break the task of expanding a local KG subgraph into multiple subtasks. We use a finite state machine with the following actions: \texttt{Think}, 
\texttt{Answer}, 
\texttt{ExpandKG}, 
\texttt{Select\_Entities}, and 
\texttt{Select\_Relation}, and states: \textit{default}, \textit{selecting-entities}, \textit{selecting-relation}, and \textit{done} as shown in Figure \ref{fig:action_state_machine}. This is named as Action State Machine (ASM) throughout the paper.

From the \textit{default} state, \tot can either \texttt{Think}, \texttt{Answer}, or choose to \texttt{ExpandKG}. After \tot chooses to \texttt{ExpandKG}, it first is prompted to \texttt{Select\_Entities} about which it needs more information (e.g., `\textit{Inception}' in Figure \ref{fig:main}). It is then prompted to \texttt{Select\_Relation} from a list of candidate relations provided by the KG interface's \texttt{get\_relations} method. After selecting a relation (e.g., `\textit{cast member}' in Figure \ref{fig:main}), all edges containing one of the selected entities as the source and the selected relation as the relation are added to the local KG subgraph. \tot is then able to \texttt{Answer}, \texttt{Think}, or \texttt{ExpandKG} again.


\paragraph{Prompt Templates.} Each state in the ASM other than the `\textit{done}' state is associated with a unique prompt template. Prompt templates are filled with information from the local KG subgraph and the KG interface before presenting them to the LLM. A customized prompt for each state allows us to present precise and relevant information along with specific instructions for each state to the LLM simplifying the LLM's task. For instance, the prompt for `\textit{selecting-entities}' presents the available options of entities to choose from that are in the local KG subgraph (see Figure \ref{fig:dynamic-prompt}). Prompt templates for all the states are in the Appendix~\ref{sec:prompt_default}-\ref{sec:prompt_sel_rel}.

Local KG subgraph is represented using a a token efficient YAML format that minimizes repetition for multiple edges on the same entity.

\begin{figure}
    \centering
    \begin{lstlisting}
Original Query: 
    |\color{CornflowerBlue}Who is Bob Dylan's maternal grandmother?|
Knowledge Graph Entities:
    |\color{ForestGreen}Q392: Bob Dylan - American singer-songwriter|
    |\color{ForestGreen}Q62519478: Beatrice Stone|
Knowledge Graph Edges:
    |\color{ForestGreen}Bob Dylan:|
        |\color{ForestGreen}mother:|
            |\color{ForestGreen}Beatrice Stone|
Previous Actions: 
...
Current task: Select entities to expand from |\color{BrickRed}[Q392,| |\color{BrickRed}Q62519478]|
Selection:
\end{lstlisting}
    \caption{\small{Prompt for the `\textit{selecting-entities}' state of the ASM. The \textcolor{CornflowerBlue}{original query} and the \textcolor{ForestGreen}{local KG subgraph} are part of the prompt for every action state while the `Current task' differs. The \textcolor{BrickRed}{entity options} show what entities can be selected based on the current local KG subgraph. If the model selected \textcolor{BrickRed}{Q62519478: Beatrice Stone} here, then in the next action state (\textit{selecting-relation}), the dynamic prompt would list the relation types that \textcolor{BrickRed}{Beatrice Stone} has edges for.}} 
    \label{fig:dynamic-prompt}
\end{figure}

\paragraph{Invoking KG Interfaces.} Outside of initialization, there are two times in which the ASM needs to invoke the KG interface: (1) When constructing the \textit{selecting-relation} prompt, the algorithm calls $\texttt{get\_relations}(\mathcal{E}_{selected})$ to gather available choices of relations. (2) When executing the transition from \textit{selecting-relation} to \textit{default} after having selected a relation type $r$, the algorithm calls $\texttt{get\_edges}(\mathcal{E}_{selected}, r)$ so that it can add the new edges and entities to the local subgraph.

\subsection{Tree Search Algorithm}
Our approach draws inspiration from the Tree-of-Thoughts approach~\cite{Yao2023TreeOT} in which an LLM is given increased reasoning power by allowing generation of multiple thoughts at a time and building a search tree over the resulting reasoning chains. We extend this approach to augment LLMs with KGs through allowing the generation of actions in addition to thoughts, and building a search tree over them. Challenges arise because Tree-of-Thoughts was not designed to incorporate actions and was not designed for knowledge intensive question answering tasks. As a result, we introduce some modifications: incorporation of actions via the ASM, a slightly different search procedure and stopping condition to better handle QA, and a different sampling procedure for improved diversity when doing constrained sampling. 

Algorithm \ref{alg:ToAction} presents the \tot tree search algorithm. Given a query $q$ it begins with initialization using $\texttt{initialize}(q)$ as described in Section~\ref{ssec:kginterface}.  After initialization, it searches by (1) choosing the unexplored node to expand based on the node's value assigned by the value function (Best First$\rightarrow$ Depth First as tie-breaker), (2) sampling $k$ actions from the LLM ($k$ is branching factor) using the prompt associated with the selected node’s state in the ASM, (3) for each of the sampled action, applying the transition function, and (4) evaluating the value of the resulting nodes with the LLM value function. The search stops when an answer is found with a node value exceeding the threshold $\tau$. To bound the search space we add two hyper-parameters: \textit{max depth} of the tree search, after which the algorithm is forced to transition to the \textit{done} state (i.e., answer the query), and \textit{max expansions} after which the model halts exploration and returns "Cannot find answer". 

\begin{algorithm}[t]
\caption{\tot}\label{alg:ToAction}
\hspace*{\algorithmicindent} \textbf{Input:} $q\gets$ query, $k\gets$ branching factor, \\\hspace*{\algorithmicindent} $\tau\gets$ answer threshold, $P\gets$ LLM,\\\hspace*{\algorithmicindent}$\mathcal{F}\gets$ ASM\\
\hspace*{\algorithmicindent} \textbf{Output:} Answer to $q$ 
\begin{algorithmic}[1]
\Procedure{ToActions}{$q$}
\State $T\gets\phi$\textcolor{Green}{\Comment{Empty tree}}
\State $Y\gets\phi$\textcolor{Green}{\Comment{Empty answer set}}
\State $s_0\gets\texttt{initialize}(q)$
\State $T$.add($s_0$)
\While {$!finished$}
\State $s\gets \texttt{choose\_node}(T)$
\\ \textcolor{Green}{\Comment{Sample $k$ actions from $P$ using the action state and associated prompt of $s$ from $\mathcal{F}(s)$}}
\State $\textbf{a}\gets \texttt{sample\_actions}(s,k,\mathcal{F},P)$ 
\For{$a\in\textbf{a}$}
\\ \textcolor{Green}{\Comment{Apply action $a$ to state $s$ according to $\mathcal{F}$}}
\State $s'\gets \texttt{transition}(s, a, \mathcal{F})$%
\\ \textcolor{Green}{\Comment{Evaluate the resulting state $s'$ using $P$}}
\State $s'.value\gets \texttt{evaluate}(s', P)$
\State $T.add(s')$
\If{$s'.action\_state=done$}
\State $Y.add(s')$
\If{$s'.value > \tau$}%
\\ \textcolor{Green}{\Comment{When answer exceeds threshold, stop search}}
\State $finished\gets True$
\EndIf
\EndIf
\EndFor
\EndWhile
\State \textbf{return} $\arg\!\max_{y\in Y}{y.value}$
\EndProcedure
\end{algorithmic}
\end{algorithm}    



\paragraph{Value Function Guidance.}
\tot computes the value of a node to determine its utility. This value is created by the LLM using evaluation prompts (step \texttt{evaluate} in Algorithm~\ref{alg:ToAction}). The value can be between 0 to 1 where 1 indicates highest utility. We use two types of evaluation prompts: one for intermediate states and one for answer states. The prompts include the original query, the local KG subgraph, the trajectory of previous actions, followed by instructions for evaluating the node (see Appendix \ref{sec:prompt_value} and \ref{sec:prompt_value_done}). These values are then used to guide the exploration of the action space. Specifically, $\texttt{choose\_node}$ returns the unexplored node with the highest value (Best First). If there are nodes with the same value, Depth First Search is used.

\paragraph{Chain-of-Traversals.} In some cases we can find the answer to a query with a single sequence of thoughts and actions using the ASM and KG interface. This is equivalent to Tree-of-Traversals with a branching factor of $k=1$. We refer to this as Chain-of-Traversals. While useful for comparison, experiments show the benefit of considering multiple branches.

\subsection{\tot with Multiple KGs}
Augmenting an LLM with more than one KG mainly involves building KG interfaces for each of the added KGs. There are a few other changes to the algorithm. (1) The entities extracted in $\texttt{initialize}(q)$ are matched with each KG interface. (2) When presenting the options of relations during \textit{selecting-relation}, $\texttt{get\_relations}(\mathcal{E}_{selected})$ is called on each KG interface. (3) When adding a new entity to the local KG subgraph we call an entity linking function for the other KG interfaces. We allow for the entity linking function to be a separate function in the KG interface or to fallback on $\texttt{initialize}(o)$ where $o$ is the text label of the entity that has just been added. This makes allows the use of explicit links between common KGs if available while still functioning without. 

\section{Experiments}
\label{sec:exp}

We evaluate \tot using three different models available on Amazon Bedrock\footnote{\url{https://aws.amazon.com/bedrock}}: Claude-Instant (\texttt{claude-instant-v1}), Llama2 70b (\texttt{llama2-70b-chat-v1}), and Llama2 13b (\texttt{llama2-13b-chat-v1}). AWS Bedrock provides on-demand access through a single API to various Foundation Models, including both open source and black box ones. This is precisely a use case \tot is designed for.

       


\begin{table*}[h]
\centering
\begin{tabular}{@{}lccccc@{}}
\toprule
           & \multicolumn{3}{c}{\textbf{2WikiMultiHop} ($\uparrow$)}    \\ \cmidrule(l){2-4}
Algorithm                                 & Claude-Instant & Llama2-70b & Llama2-13b \\ \midrule
\multicolumn{1}{l|}{Chain-of-Thought}     & 25.2           & 30.4       &  26.8        \\
\multicolumn{1}{l|}{ReAct}               & 5.8            & 30.4       &  6.8     \\
\multicolumn{1}{l|}{ReAct$\rightarrow$ CoT}& 27.0 (84.8\%) & 41.6 (41.6\%)& 23.6 (71.8\%)        \\ 
\multicolumn{1}{l|}{FLARe}               & 35.0  & 45.4        &  34.8  \\ 
\multicolumn{1}{l|}{Chain-of-Traversals} & 42.6           & 45.0       &  31.0        \\
\multicolumn{1}{l|}{\textbf{Tree-of-Traversals}}  & \textbf{63.0} & \textbf{56.6} & \textbf{37.3} \\ \midrule
& \multicolumn{3}{c}{\textbf{QALD-10} ($\uparrow$)}   \\ \cmidrule(l){2-4} 
Algorithm                                & Claude-Instant & Llama2-70b & Llama2-13b \\ \midrule
\multicolumn{1}{l|}{Chain-of-Thought}    & 47.4           & 40.0       &   \textbf{42.6}      \\
\multicolumn{1}{l|}{ReAct}               & 3.1            & 23.7       & 8.7   \\
\multicolumn{1}{l|}{ReAct$\rightarrow$ CoT}& 47.9 (72.6\%)  & 43.4 (44.4\%)&  40.8 (79.5\%)  \\
\multicolumn{1}{l|}{FLARe}               & 39.2           & 38.8      & 23.1   \\
\multicolumn{1}{l|}{Chain-of-Traversals} & 55.7           & 56.9       &  39.6        \\
\multicolumn{1}{l|}{\textbf{Tree-of-Traversals}}  & \textbf{64.3} & \textbf{61.6} & 39.2 \\ \bottomrule
\end{tabular}
\caption{EM-in on 2WikiMultiHop dataset. $\rightarrow$CoT indicates falling back to Chain-of-Thought when no answer is given. (\%) indicates the number of such cases.}
\label{tab:2wiki}
\end{table*}

\subsection{Tasks and Datasets.} We first evaluate the \tot algorithm on two common tasks used for evaluating an LLM's knowledge: 2WikiMultiHop and QALD-10. To allow testing on complex questions requiring knowledge from multiple KGs we create a new dataset that requires knowledge from multiple KGs.

\textbf{2WikiMultiHop Dataset}~\cite{xanh2020_2wikimultihop} was constructed by extracting multi-hop templates from HotPotQA~\cite{Yang2018HotpotQAAD}, combining templates to get complex reasoning questions, generating candidate questions with Wikidata, and then confirming that the entity mentions for each edge appear in the Wikipedia paragraphs. Answers to these questions can be derived from both Wikipedia and Wikidata~\cite{Vrandei2014WikidataAF}. We subsample 500 questions from the test set following the approach used in ReAct~\cite{Yao2022ReActSR}, including the sampling seed value of 233.

\textbf{QALD-10 Dataset}~\cite{usbeckqald}  is a multi-lingual Knowledge Graph Question Answering (KGQA) dataset with 395 questions created by humans, translated into other languages, and then constructed as a SPARQL query over Wikidata\footnote{\url{https://github.com/KGQA/QALD-10}}. These questions are more varied in terms of reasoning structure than the questions in 2WikiMultiHop (e.g. requiring multiple answers or aggregations). We used the questions in English language. 

\begin{figure}
    \centering
    \begin{tikzpicture}
        \node [draw, align=center] (title) at (0,0) {Which event was \textcolor{Mahogany}{hosted in} a venue with a larger\\\textcolor{ForestGreen}{maximum capacity}, \textcolor{Mahogany}{\textit{Fearless Tour: Indianapolis}}\\or \textcolor{Mahogany}{\textit{Fearless Tour: Omaha}}?};
        
        \node[inner sep=0pt] (mb) at (3.3, -1.3)
        {
        \includegraphics[width=0.04\textwidth]{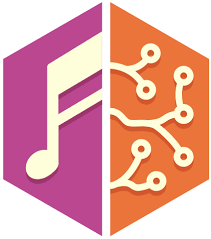}
        };

        \node[inner sep=0pt] (mb) at (3.3, -2.3)
        {
        \includegraphics[width=0.04\textwidth]{figures/musicbrainz_logo.png}
        };

        \node[inner sep=0pt] (mb) at (3.3, -3.3)
        {
        \includegraphics[width=0.06\textwidth]{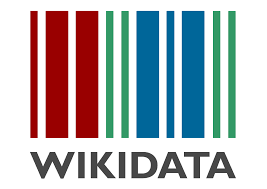}
        };

        \node[draw, overlay, align=left, text width=7.45cm] at (0,-2.8)
        {
        1. Get \textcolor{Mahogany}{venue} for \textcolor{Mahogany}{\textit{Fearless Tour: Indiana-\\\ \ \ \ polis}} $\rightarrow$ \textcolor{CornflowerBlue}{\textit{Gainbridge Fieldhouse}}\\
        2. Get \textcolor{Mahogany}{venue} for \textcolor{Mahogany}{\textit{Fearless Tour: Omaha}}\\\ \ \ \ $\rightarrow$ \textcolor{CornflowerBlue}{\textit{CHI Health Center Omaha}}\\
        3. Get \textcolor{ForestGreen}{maximum capacity} of the respec-\\\ \ \ \ tive venues $\rightarrow$ (18165, 18975)\\
        4. Answer: \textit{Fearless Tour: Omaha was hosted in a venue with a larger maximum capacity.}
        };

        \node[] at (0,-4.4) {};

    \end{tikzpicture}
    \caption{Example question from MusicBrainz-x-Wiki and how \tot arrives at its final answer. \textcolor{Mahogany}{Red} indicates entities and and relationships belonging to MusicBrainz. \textcolor{ForestGreen}{Green} indicates relationships only in Wikidata. \textcolor{CornflowerBlue}{Blue} indicates entities linked to both KGs.}
\end{figure}
\label{fig:cross-kg}

\textbf{MusicBrainz-x-Wikidata Dataset} One novel use-case of \tot is synthesizing and reasoning over multiple knowledge graph sources. There is no existing dataset that requires synthesizing information from multiple KGs to answer individual questions. Therefore, to test the reasoning ability using multiple KGs, we create a new dataset, MusicBrainz-x-Wikidata, containing 109 questions that require reasoning with information from both MusicBrainz and Wikidata. Unlike Wikidata which contains general knowledge, MusicBrainz is a deep domain-specific database on the music industry. The vast majority of this information is unlikely to be known by large language models. We construct the MusicBrainz-x-Wikidata dataset with human annotators who were provided with instructions to find reasoning paths between these two KGs, the question types to focus on, and some example questions. The curated questions were checked for ambiguity and sensibility. By design, each question requires extracting information from both knowledge graphs in order to answer it successfully. In addition to the reasoning types in 2WikiMultiHop, this dataset  contains questions that involve aggregations, comparisons of aggregations, qualifications, and complex combinations of these. An example can be seen in Figure \ref{fig:cross-kg}. Detail on instructions, question types, and examples are in Appendix~\ref{sec:app_music_x_wiki}. 

\subsection{Metric.} We use Exact Match Included (EM-in) as the evaluation metric. EM-in is 1 if the ground truth answer appears with an exact match anywhere in the answer and 0 otherwise. This accounts for the propensity of an LLM to output answers in a sentence with varying syntax. This is a common metric but is often referred to interchangeably with Exact Match (EM) \cite{Sun2023ThinkonGraphDA}. When there are multiple answers, we compute average EM-in over all labels. 

\subsection{Comparison Baselines.} We experiment against three related approaches that can work with any black-box LLMs: (1) Chain-of-Thought (CoT) prompting~\cite{Wei2022ChainOT} (2) ReAct~\cite{Yao2022ReActSR} which iterates between generating thoughts and generating actions for searching and retrieving from a text knowledge base like Wikipedia, and (3) Forward Looking Active Retrieval (FLARe)~\cite{Jiang2023ActiveRA} which iterates between generating thoughts and then retrieving from a knowledge base to correct inaccuracies. 

\subsection{Implementation details.} For all models, we use a sampling temperature of 0.0 for the LLM when multiple samples are not required and a temperature of 1.0 when diverse samples are required. We test our approach in two settings: (1) with a branching factor of $k=1$ termed as Chain-of-Traversals, and (2) with a branching factor of $k=3$ termed as Tree-of-Traversals. In both setups, we set maximum depth to 7 which means that upon reaching the \textit{default} action state beyond depth 7, the only available action is to answer the question. For Tree-of-Traversals, we set the maximum total expansions to 20. The answer threshold $\tau$ is set to 0.8 which corresponds to high confidence answers supported by the KG according the $\texttt{evaluate}$ prompt (Appendix \ref{sec:prompt_value_done}).


For 2WikiMultiHop and QALD-10, we use Wikidata as the knowledge graph~\cite{Vrandei2014WikidataAF}. 
For MusicBrainz-x-Wikidata, we use MusicBrainz in addition to Wikidata. 
We implement the KG interface for Wikidata using Wikidata SPARQL queries\footnote{\url{https://query.wikidata.org/}}, and implement the KG interface for MusicBrainz KG using the MusicBrainz API\footnote{\url{https://musicbrainz.org/doc/MusicBrainz_API}}.

\section{Results and Discussion}

\begin{table*}[h]
\centering
\begin{tabular}{@{}lccc@{}}
\toprule
           & \multicolumn{3}{c}{\textbf{Music-x-Wiki} ($\uparrow$)}   \\ \cmidrule(l){2-4} 
Algorithm  & Claude-Instant & Llama2-70b & Llama2-13b \\ \midrule
\multicolumn{1}{l|}{Chain-of-Thoughts}   & 11.9           & 10.1       &  13.8        \\
\multicolumn{1}{l|}{Chain-of-Traversals} & 22.0           & 21.9       & 14.7   \\
\multicolumn{1}{l|}{\textbf{Tree-of-Traversals}} & \textbf{40.4} & \textbf{23.4} & \textbf{16.5} \\ \bottomrule
\end{tabular}
\caption{EM-in on MusicBrainz-x-Wikidata dataset.}
\label{tab:cross_kg}
\end{table*}

Table~\ref{tab:2wiki} presents the results of our experiments on 2WikiMultiHop and QALD-10. For all the models, \tot outperforms the baselines on 2WikiMultiHop, setting state-of-the-art results for these tasks in the zero-shot setting. We hypothesize that much of this gain is due to Tree-of-Traversals' access to the knowledge base via proposed KG interface and its thought-action procedure guided by the ASM. This is evident, as even Chain-of-Traversals, which does not perform a tree traversal (including multiple thoughts/actions, backtracking, and node value computation), significantly outperforms ReAct's knowledge grounding: 8.7\% higher accuracy than ReAct$\rightarrow CoT$ on 2WikiMultiHop when averaged over all models. Compared to Chain-of-Traversals, \tot further improves performance. It gives on average a 12.8\% absolute accuracy increase for 2WikiMultiHop and 4.3\% absolute improvement for QALD-10. We note that the better performing the model is, the more it stands to gain from \tot as noted by the difference between Llama-70b and Llama-13b. 

\begin{figure}
    \centering
    \includegraphics[width=\linewidth]{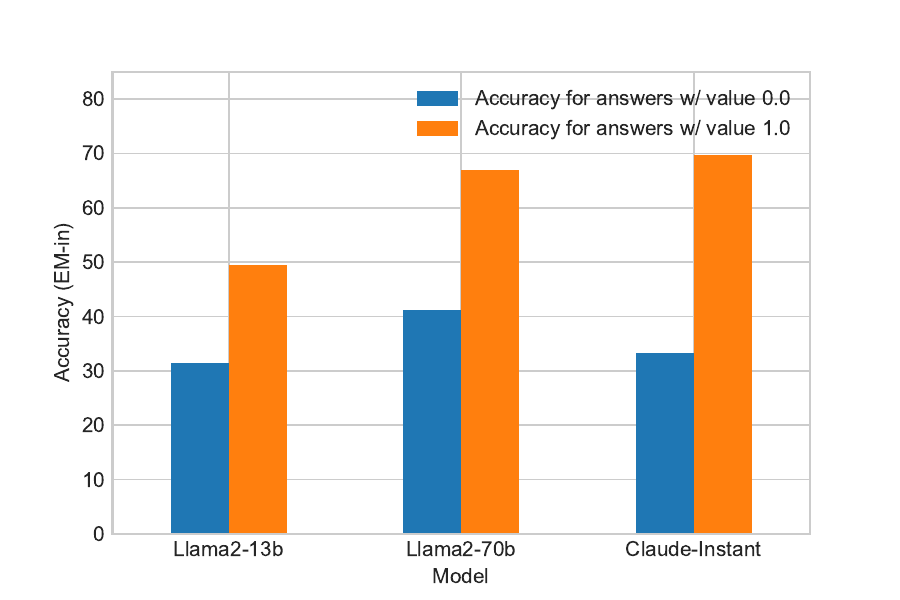}
    \caption{\small{The EM-in accuracy for answers with model assigned value 0.0 or 1.0 and the corresponding true EM-in score of the answer. This includes all proposed answers, not just the final answer returned by the model.}}
    \label{fig:ans_value_correctness}
\end{figure}

On MusicBrainz-x-Wikidata (Table~\ref{tab:cross_kg}), which contains challenging reasoning questions requiring access to two KGs, there is an average of 37.4\% relative improvement from \tot over Chain-of-Traversals as shown in Table~\ref{tab:cross_kg}. Both Chain-of-traversals and \tot outperform Chain-of-Thoughts on this dataset. We only compare with Chain-of-Thoughts as other retrieval methods have low coverage over the MusicBrainz knowledge base. 


\subsection{Effect of the Value Function.} \tot relies on signal from the value function to pick the final tree trajectory. If the values were arbitrary, then Tree-of-Traversals would not do any better than Chain-of-Traversals. Figure~\ref{fig:ans_value_correctness} shows that there is a meaningful signal from the value function for all models. There is an average performance difference of 31.0\% between the accuracy for answers valued at 1.0 vs 0.0. This represents an average relative improvement of 83.2\% when selecting answers valued at 1.0 over those valued at 0.0. See Appendix \ref{sec:value_function_profiles} for individual profiles of each model's value function.

\subsection{Effect of Backtracking.} 
\begin{figure}
    \centering
    \includegraphics[width=\linewidth]{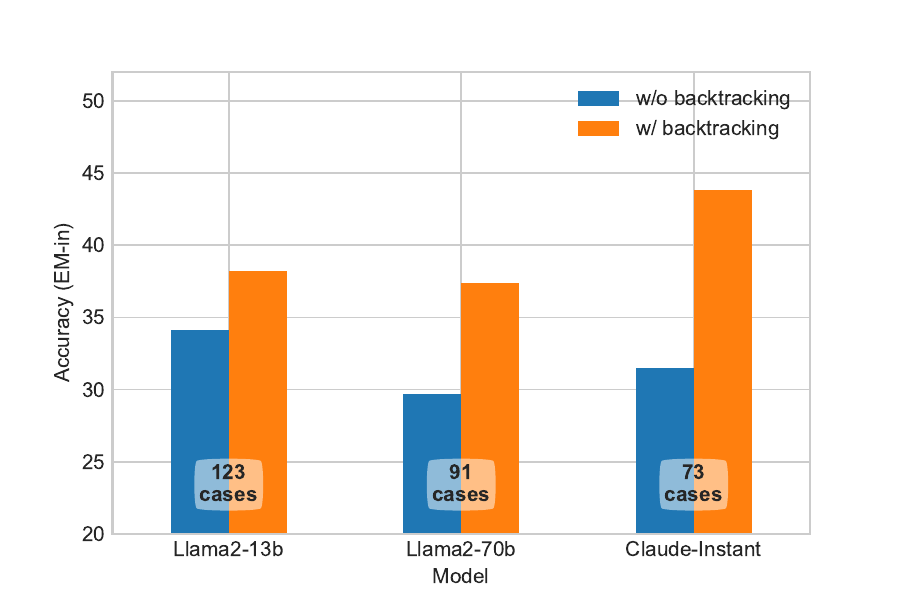}
    \caption{\small{Comparison of results with and without backtracking on 2WikiMultiHop questions that needed backtracking. There would be a major degradation in performance if backtracking was not allowed.}}
    \label{fig:backtracking}
\end{figure}

To determine the effect of backtracking in Tree-of-Traversals, we ask the counter-factual of how would the model perform if it could not backtrack (Figure \ref{fig:backtracking}). We limit the analysis to 2WikiMultiHop questions in which Tree-of-Traversals generates answers on a subtree, and the model eventually backtracks on that subtree (i.e., the cases where we have a counter-factual). As a result, these questions are generally more challenging than the overall distribution of questions. In these cases, we compare the result if the highest-valued answer was taken from the first searched subtree compared to the ultimate answer after backtracking. We find that the ability to backtrack gives \tot a significant accuracy increase ranging from 4.1\% to 12.3\%.  

\subsection{Performance on MusicBrainz-x-Wikidata.} For MusicBrainz-x-Wikidata dataset, we only compare against Chain-of-Thought since the implementations for other baseline algorithms do not have access to similar music-specific knowledge bases. Despite this, we observed some interesting results. Chain-of-Thought does far worse on MusicBrainz-x-Wikidata than on the more general datasets based exclusively on Wikipedia/Wikidata (10.1\%-13.8\% vs. 25.2\%-47.4\%). This could be because LLMs are trained on vast amount of general knowledge such as that found in Wikipedia, but they are not trained with as much data from specific domains such as music. Besides the presence of KG interface and ASM with Tree-of-Traversals, this could be an additional reason why Tree-of-Traversals does 2.2 times as well as Chain-of-Thought on MusicBrainz-x-Wikidata. This demonstrates the importance of augmenting LLMs with domain-specific KGs and/or multiple KGs which the proposed Tree-of-Traversals is capable of.

\subsection{Analysis of Baselines.} ReAct underperforms Chain-of-Thought in most cases. This phenomenon was seen in the original ReAct paper which noted that their approach significantly reduced hallucinations despite lower accuracy~\cite{Yao2022ReActSR}. Therefore, falling back on Chain-of-Thought results in an accuracy improvements for for Llama2-70b and Claude-Instant. We note that \texttt{claude-instant} vastly underperforms Llama2-70b on ReAct. The primary reason for this is \texttt{claude-instant} refuses to "search" people and apologizes after performing invalid actions. Despite this, ReAct$\rightarrow$CoT still improves over CoT on both 2WikiMultiHop and QALD-10. FLARe improves model performance on 2WikiMultiHop but not on QALD-10. This may be due to better overlap between the text knowledge base for 2WikiMultiHop. 

\section{Related Works}
\label{sec:rel_works}
\paragraph{Knowledge Base Question Answering} There is a large history of work that has looked into answering questions using a knowledge graph \cite{Wu2019ASO, Lan2021ASO}. The approaches can broadly be broken into those that attempt to parse the question into a logical form \cite{Berant2014SemanticPV, Luo2018KnowledgeBQ, Zhu2022NeuralSymbolicMF}, and information retrieval approaches \cite{Bordes2015LargescaleSQ, Chen2019UHopAU}

\paragraph{Knowledge Enhancement.} Recently, researchers have looked into enhancing pretrained language models and LLMs using knowledge graphs. Many works have focused on incorporating knowledge graph data into LLMs through training, often with architectural changes to the model~\cite{Zhang2019ERNIEEL, Wang2019KEPLERAU, Peters2019KnowledgeEC, Yamada2020LUKEDC, He2021ImprovingMK}. Some of these have found success using specialized graph encoder layers \cite{Yasunaga2021QAGNNRW, Sun2021ERNIE3L}. Some have sought to mix this process with pretraining of the language model \cite{Yasunaga2022DeepBL}. The limitations of including KGs during LLM training are: the models are incapable of incorporating KG updates without retraining, are not able to change KG source (e.g., a domain-specific one), and may have low explainability resulting from learning knowledge in model weights rather than explicit retrieval. In addition, these methods add complexity to the training process, increase cost, and do not work with LLM without access to model weights. As a result, scaling these methods has often been seen as too risky. 

Other methods treat a KG as a more complete source of truth and use the LLM to generate structured queries for the KG. \citet{Tianle2023FewshotIL} and \citet{Choudhary2023ComplexLR} teach the LLM to generate a logical KG query which then returns matching entities from the KG. These methods share similarities with the semantic parsing based approaches mentioned earlier. However, by returning a logical query for the KG, these methods lose the reasoning and commonsense abilities of the LLM. 

Some approaches have looked at injecting KG or knowledge base data into LLM prompts. Many methods do a single round of retrieval based on the query~\cite{Lewis2020RetrievalAugmentedGF, LI2023GraphRF} using a retrieval mechanism, such as dense passage retrieval~\cite{Karpukhin2020DensePR}. These methods cannot answer more complex multi-hop questions as the initial retrieval is unlikely to contain the secondary or tertiary information that will ultimately be required. Later methods have made the retrieval process iterative. E.g. FLARe~\cite{Jiang2023ActiveRA} uses prediction of the upcoming sentence to retrieve relevant documents and regenerates until the sentence contains high-confidence tokens. In \citet{Wang2023KnowledgeDrivenCE} multi-round QA format is used for multi-round retrieval. ReAct \cite{Yao2022ReActSR} does multiple rounds of thoughts and actions to query a text-based knowledge base. 
\citet{Jiang2023StructGPTAG} repeatedly interfaces with a KG. In other parallel works, \cite{Sun2023ThinkonGraphDA, Wen2023MindMapKG} explore methods for building KG context for LLMs using path and neighborhood based search methods. The above methods do not incorporate tree search or forms of advanced reasoning beyond the LLM's innate ability. They cannot explore multiple reasoning paths or solutions and cannot backtrack if making a mistake. This results in lesser capabilities. Additionally, none of the above methods explore reasoning over multiple KGs. 

\paragraph{Multiple Knowledge Base QA.} Some works studied QA on questions derived from differing domains. These works learn to pick between different domain-specific QA models, each trained on a single knowledge base \cite{Puerto2021MetaQACE,Geigle2021TWEACTW,Puerto2023UKPSQuAREVA}. Besides requiring training, such systems cannot answer questions requiring the synthesis of information from different domains (MusicBrainz-x-Wikidata). 

\section{Conclusion}

Tree-of-Traversals is a powerful algorithm that enables LLMs to utilize KG reasoning with zero examples, no schema dependency, and no training. We demonstrate its efficacy experimentally on multiple LLMs and datasets. We hope \tot continues to be developed and see value in studying integration with personalized user KGs as well as with other domain-specific datasets. 

\section{Limitations}
\label{sec:limitations}
Tree-of-Traversals is slower than simpler retrieval alternatives. Improving the value function would reduce the number of LLM and KG API calls by avoiding incorrect paths along the tree to explore. Other engineering solutions, such as hosting graph servers, could be implemented to accelerate LLM and KG access. In terms of the token cost, the KG text representation is token efficient compared to many retrieval methods using unstructured knowledge bases as the latter tend to maximize usage of the LLM context window. However, compared to non-retrieval baselines, there is significant increase in token usage cost. 

The types of KG questions that can be answered are limited by the context window, search depth, and LLM's reasoning ability. For instance, a very large aggregation (`\textit{How many mountains have elevation above 3500 meters?}') would require more entities than what fits in the LLM context. Future work could look at adding other actions to the ASM, such as aggregations, that could distill the local KG into more relevant information. 

EM-in is also an imperfect metric. False negatives can arise from discrepancies in date and number formatting, text formatting, and aliasing. False positives can arise in cases in which the model does not definitively answer but includes the correct answer in its response.

\section{Ethical Impact}

We do not anticipate \tot to introduce new areas of risk but it may have unstudied effects on existing risks of LLMs or KGs. We highlight the following areas. (i) \tot gives new capabilities to LLMs after training. While positive in terms of accuracy, we have not evaluated its effect on wider safety metrics. (ii) Our evaluation is limited to only English versions of KGs and datasets. \tot should be evaluated in other languages to ensure consistent and fair experience. (iii) We have not performed analysis under misleading or deceptive knowledge graphs. Using a publicly modifiable knowledge graph does come with the risk that information could be deceptively changed.

In terms of positive ethical impact, making this research public democratizes access to knowledge-augmented LLMs as this method does not require a large training investment or owning the language model to build and customize. 

\bibliography{tree_of_traversals}
\bibliographystyle{acl_natbib}

\appendix

\section{MusicBrainz x Wiki}
\label{sec:app_music_x_wiki}

\begin{table*}[h!]
\begin{tabular}{| p{0.15\textwidth} | p{0.06\textwidth} | p{0.32\textwidth} | p{0.36\textwidth} |}
\hline
\textbf{Question Type }                              & \textbf{Count} & \textbf{Description }                                                                                            & \textbf{Example}                                                                                                                       \\ \hline
Bridge                                      & 30    & Direct multi-hop questions                                                                              & The father of Alexander Newley was the vocal arranger for what song?                                                          \\ \hline
Bridge Comparison                           & 7     & Comparison of values from two multi-hop reasoning chains                                                & Which venue was opened more recently, the venue for The Eras Tour: Seattle (night 1) or the venue for Fearless Tour: Seattle? \\ \hline
Bridge Aggregation                          & 42    & Count of the number of entities satisfying a multi-hop condition (aggregation only over the second hop) & The place which is the filming location of Michelangelo Buonarroti, has how many songs recorded there?                        \\ \hline
Bridge Aggregation Comparison               & 7     & Comparison of counts of entities between two different aggregation conditions & Who composed more songs, Marie Daulne's mother or Qubilah Shabazz's godparent?                                                \\ \hline
Qualification Bridge                        & 10    & Multi-hop question in which one of the edges needs to be disambiguated with a qualifying statement      & The filming location of Scream Awards that was built in 1929, has what recording engineered there?                            \\ \hline
Qualification Bridge Aggregation            & 9     & Count of the number of entities satisfying a qualification bridge style condition                       & The sibling of Teppo Ruohonen that was born in 1949, has composed how many songs?                                             \\ \hline
Qualification Bridge Aggregation Comparison & 1     & Comparison of qualification bridge aggregation type questions                                           & Which of the children of Isabel Preysler that are singers has released more albums?                                           \\ \hline
Double Qualification                        & 3     & Multi-hop question in which each hop requires a disambiguating qualification edge                       & The child of Guus Belinfante that was a singer, has a vocal feature on what recording by Doe   Maar?                          \\ \hline
\end{tabular}
\caption{Composition of questions for Musicbrainz x Wikidata}
\label{tab:music-x-wiki-questions}
\end{table*}

Additional information on the creation of the MusicBrainz x Wiki dataset.

\subsection{Creation steps}

The dataset was created by annotators in a semi-automatically supported manner. We developed an automated tool to support the following process: 
\begin{itemize}
    \item Find linking entities that are present in both MusicBrainz and Wikidata of various types (artist, label, place, event, etc.)
    \item Find related entities that could be used in a question to perfectly identify the entities above (e.g., The place owned by the Detroit Tigers $\rightarrow$ Tiger Stadium)
    \item Find related entities that could be used in a question to ambiguously identify a linking entity, so that a qualifier could be added to disambiguate the linking entity.
    \item Count the number of edges in Wikidata or MusicBrainz for each relationship type an entity has.
\end{itemize}

We use these tools to create an initial set of multi-hop questions falling into specific reasoning categories. We then have human annotators check each question for sensibility (grammatically correct and can be definitively understood) and ambiguity (one right answer in the KGs). If possible, the question is reworded or fixed to not contain ambiguities. We are then left with a final set of 109 questions. 

\subsection{Composition of Questions}

The composition of questions for Musicbrainz x Wiki can be found in table \ref{tab:music-x-wiki-questions}.

\subsection{Annotators}

The annotators consisted of approximately 10 research scientists and engineers, all English speaking and based in the United States.

\section{Implementation Details}

\subsection{Diversity Oversampling.}
One problem with Tree-of-Thoughts is that when there are constrained options to choose from, the LLM becomes repetitive and fails to produce diverse outputs~\cite{Yao2023TreeOT}. Their solution used a "propose prompt" with additional instructions for proposing multiple distinct thoughts. However, this adds significant complexity to the prompt for purposes that are only tangential to the task being completed. A simpler approach we employ is to generate diverse actions by oversampling. Specifically, if the branching factor is $k$, we sample $2*k$ possible actions from the LLM and then extract the first $k$ unique ones to use. We only employ this for \textit{selecting-entities} and \textit{selecting-relation} as the action choices are limited and diversity is required. This change does not noticeably affect the computation cost of \tot as multiple actions are sampled for the same prompt and the average prompt input size (100s-1000s of tokens) is generally much larger than the average generation length for these actions (<20 tokens). Latency remains similar as multiple actions are sampled in parallel. With this, we consistently generate diverse options when \textit{selecting-entities} or \textit{selecting-relation}. 

\subsection{Hyperparameters}

\tot does not have a huge number of hyperparamters. Most of the hyperparameters were chosen analytically, though some were chosen based on preliminary experiments. We share some information on choosing them.

$\tau$ was chosen analytically in conjunction with the prompt for \texttt{evaluate} on the \textit{done} state (Appendix \ref{sec:prompt_value_done}). Chosen to be 0.8 so that \tot only stops when it is confident that the answer is supported by the KG. 

The temperature was chosen to be 1.0 to promote diversity. If implementing with other models which can have temperature >1.0, we recommend hyperparameter tuning or using a sample prompt of each type and analytically choosing the temperature such that diverse but sensible actions are output. 

The branching factor $k=3$ was chosen based on small scale experimentation. Larger $k$ makes the model spend longer searching, taking longer time and increasing expenses. Smaller $k$ did not generate as much diversity and options. It is worth noting that even with $k=3$ and diversity sampling, sometimes the actual number of unique outputs is less than 3. 

The max depth was chosen analytically to be minimal while enabling answering of most questions. Some questions in QALD-10 and MusicBrainz-x-Wikidata would require a greater depth to answer. 

The maximum expansions cutoff was chosen purely to ensure the model does not hang on a single question too long. The actual value is somewhat arbitrary and has minimal impact on accuracy.

\subsection{KG Interface}

We present some additional details of the KG Interface. 

The \texttt{initialize} function is a three step process. First a LLM call and prompt is used to extract named entities. Second, for each extracted entity candidates are searched for using the KG API. Finally, another LLM call is used to match the extracted entity to the best candidate. 

\texttt{get\_relations} and \texttt{get\_edges} are implemented exclusively using the respective KG API. For Wikidata, these are each one or two SPARQL query (forward and reverse edges). For MusicBrainz, they each require multiple API calls as the endpoints are separated for different entity types. E.g. there is a separate endpoint for Artists and Recordings.

\section{Value Function Profiles}
\label{sec:value_function_profiles}

\begin{figure}
  \centering
  \subfloat[][Llama2-13b]{\includegraphics[width=.4\textwidth]{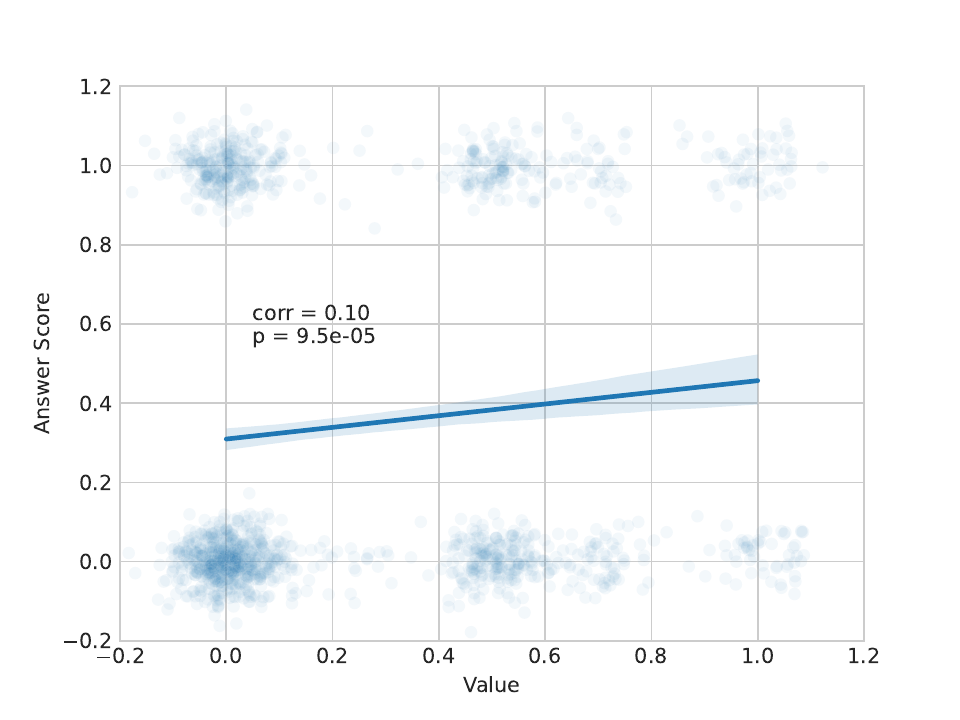}}\\
  \subfloat[][Llama2-70b]{\includegraphics[width=.4\textwidth]{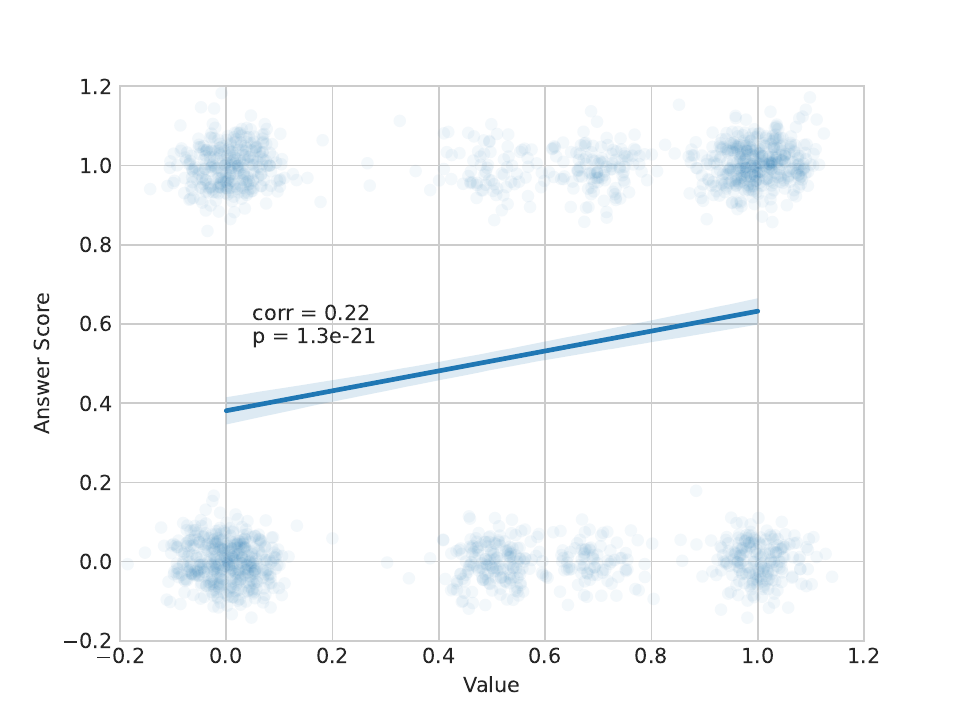}}\\
  \subfloat[][claude-instant]{\includegraphics[width=.4\textwidth]{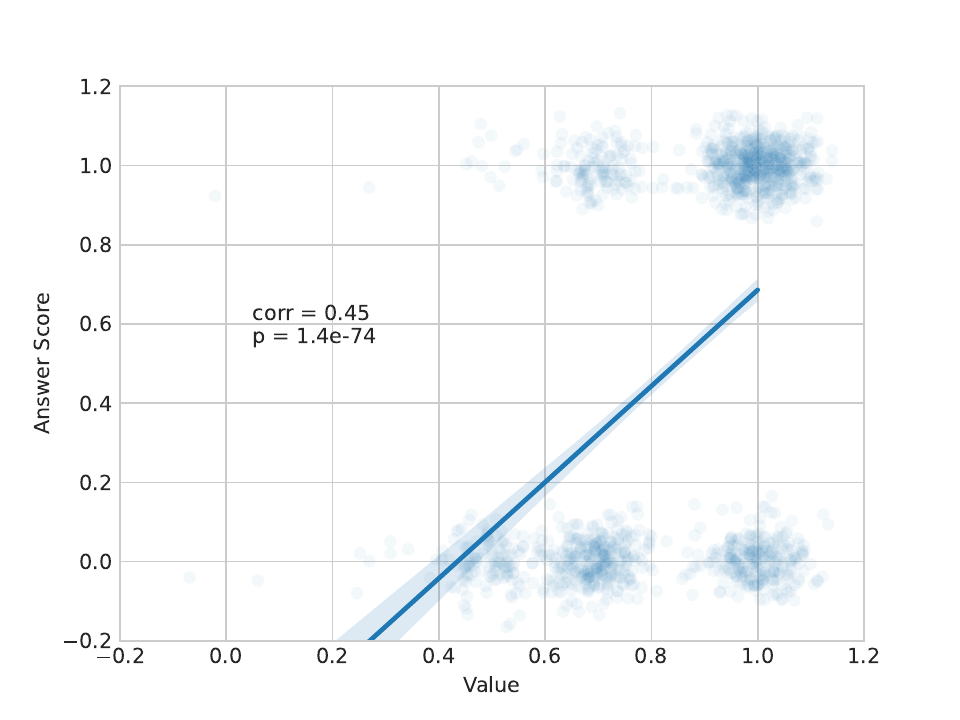}}\\
  \caption{EM-in Accuracy vs value function for all \tot answers on 2Wiki. Points are jittered for visibility. Trendline indicates correlation and p-value. }
  \label{fig:value_acc_grid}
\end{figure}

We profile each value model's value function to assess their characteristics. Figure \ref{fig:value_acc_grid} shows the individual answer value vs accuracy profiles for each model on 2WikiMultiHop. All models have a positive correlation between the answer value and the answer's accuracy. However, as is to be expected, there are differences with Llama2-13b having the lowest correlation and Claude-Instant having the highest correlation. 

\begin{figure}
  \centering
  \subfloat[][Llama2-13b]{\includegraphics[width=.4\textwidth]{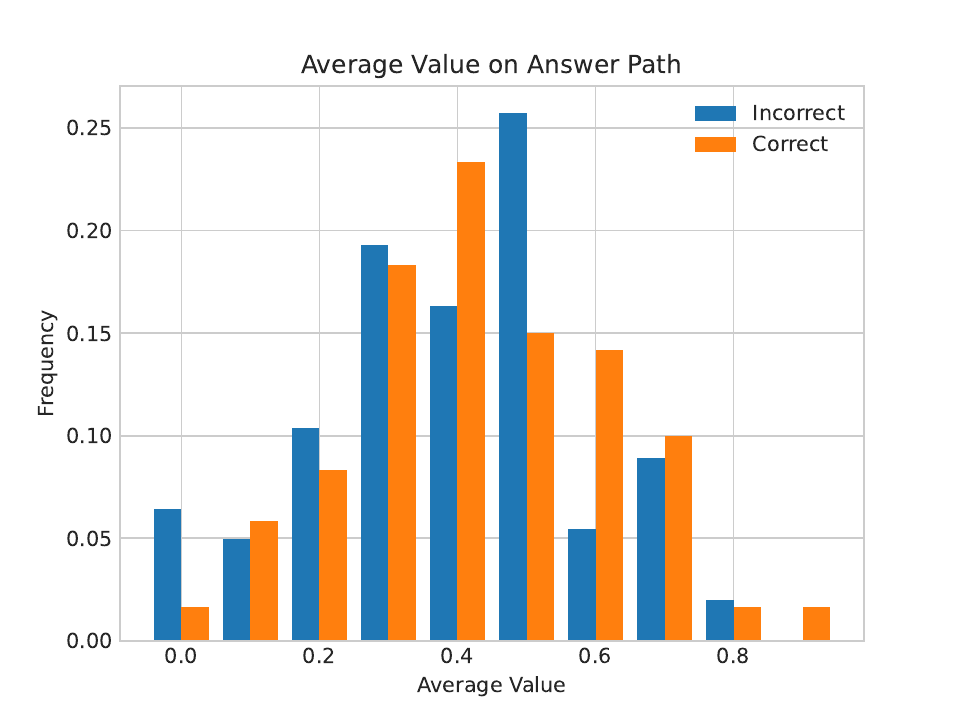}}\\
  \subfloat[][Llama2-70b]{\includegraphics[width=.4\textwidth]{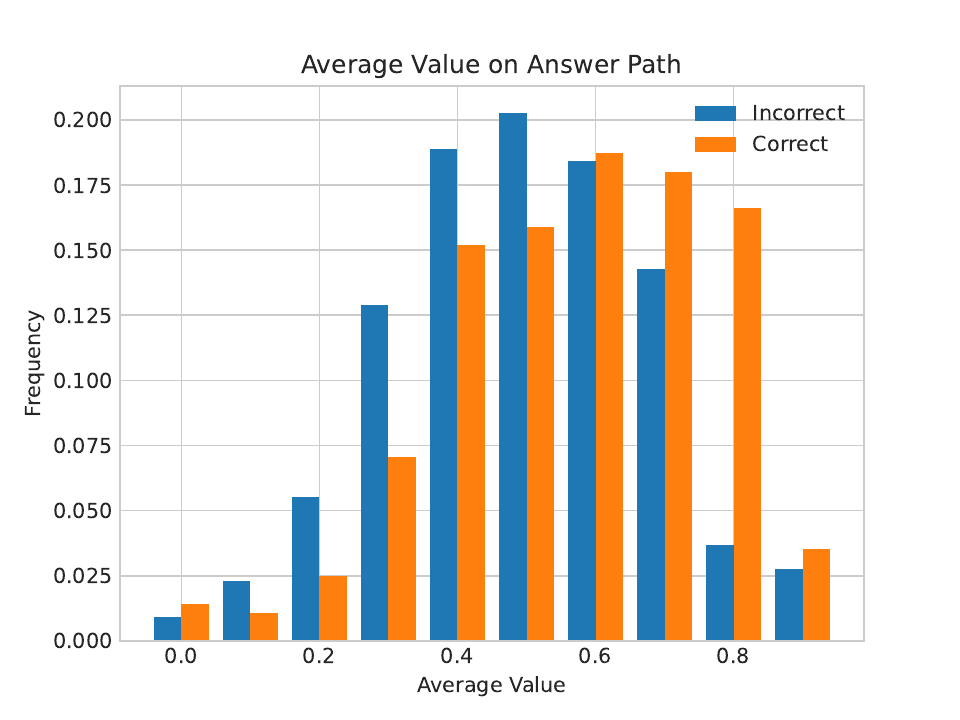}}\\
  \subfloat[][claude-instant]{\includegraphics[width=.4\textwidth]{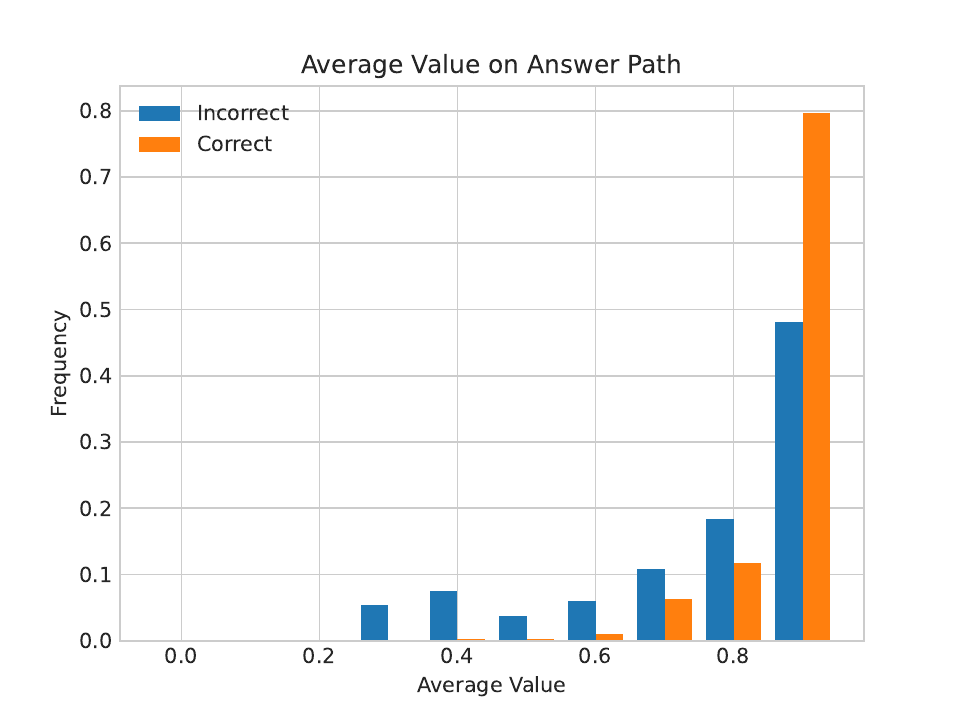}}\\
  \caption{The average value function of intermediary steps along the path to each answer, separated by whether the answer would be scored as correct or incorrect. For each answer proposed by the model, the scores along the path are averaged, and the resulting distribution plotted. Data is from 2WikiMultiHop results.}
  \label{fig:value_on_path}
\end{figure}

Figure \ref{fig:value_acc_grid} only look at the value function response for final answer states as those are the only ones with a directly associated accuracy score. While accurately evaluating answer states is more important, we also want to see some useful signal from the values for intermediary states. Figure \ref{fig:value_on_path} shows the average value for nodes on the search path to the answer, separated by whether the answer scores as correct or not. We would expect actions that ultimately lead to a correct answer to have higher values than those that lead to incorrect ones. We clearly see this with the \textit{correct} distribution being further right than the \textit{incorrect} distribution. This is least pronounced for Llama2-13b which is to be expected. We also note that the distributions shape is different between the Llama2 models and Claude-Instant. 

These results indicate that larger, stronger models, will likely see more improvement in the value function, and thus, more utility from \tot. 

\section{Prompt Engineering}
\label{sec:prompt_engineering}

Some prompt re-engineering was required when switching between models for both our methods and ReAct. This engineering was limited primarily to formatting of responses. For example, Llama2 would start responses with "So the next action should be" while we required the action to start with "THINK", "EXPAND\_KG" or "ANSWER". Instead we incorporated ``So the next action should be:'' into the prompt so that the output began with the chosen action. For \texttt{selecting-entities} we added ``You have selected the following entities to expand:''. For \texttt{selecting-property} we added ``I suggest selecting property:''. These issues could also be solved by changing the parsing of the presented prompts. 

\section{Fallback to Chain-of-Thought ($\rightarrow$CoT)}

We fallback to Chain-of-Thought for the ReAct baseline when either no answer is provided after depth 7 or when one of the following phrases appears in the answer: 'determine', 'unable', 'cannot', 'unknown', 'unsure', or 'not possible'. While these certainly could appear in intentional answers, we find that they occur consistently and exclusively in non-answer style responses.

\onecolumn
\section{Prompts}

The following pages contains the prompts used in Tree-of-Traversals. The dynamic components that change based on the query, KG state, action history are shown in color.

\subsection{Prompt for \textit{default} action state}
\label{sec:prompt_default}
\begin{minipage}{\linewidth}
        \begin{lstlisting}
Original Query: 
    |\color{CornflowerBlue}Who is Bob Dylan's maternal grandmother?|
    
Knowledge Graph Entities:
    |\color{ForestGreen}Q392: Bob Dylan - American singer-songwriter|
    |\color{ForestGreen}Q62519478: Beatrice Stone|
Knowledge Graph Edges:
    |\color{ForestGreen}Bob Dylan:|
        |\color{ForestGreen}mother:|
            |\color{ForestGreen}Beatrice Stone|

Previous Actions: 
|\color{ForestGreen}EXPAND KG: I should search for the mother of Bob Dylan|
|\color{ForestGreen}SELECT ENTITIES: Q392|
|\color{ForestGreen}SELECT RELATION: P25 - has mother|

You are a superintelligent AI equipped with the ability to search a knowledge graph for definitive, up-to-date answers. Your task is to interface with the knowledge graph in order to answer the above query. You will be able to expand the knowledge graph until you have found the answer. Think in detail before acting or answering.

Available actions:
'THINK' - Generate relevant thoughts to solving the problem. This could include recalling well known facts from memory.
    e.g.,
        THINK: I should search for the movies directed by...
        THINK: I know that Biden is American, therefore...
        THINK: I see that John Cena is in both The Independent and Vacation Friends, therefore...
'EXPAND_KG' - Search for edges of entities in the knowledge graph using an external API. This is a useful function for getting to the correct answer.
    e.g., EXPAND_KG: I should search for the country of origin of Jonathon Taylor
'ANSWER' - Generate the final answer once the problem is solved. Just state the best answer, do not output a full sentence.
    e.g., 
        ANSWER: No
        ANSWER: Harry Potter
        ANSWER: [Harry Potter, Ron Weasley, Hermione Granger]
\end{lstlisting}
\end{minipage}

\subsection{Prompt for \textit{selecting-entities} action state}
\label{sec:prompt_sel_ent}

\begin{minipage}{\linewidth}
        \begin{lstlisting}
Original Query: 
    |\color{CornflowerBlue}Who is Bob Dylan's maternal grandmother?|
    
Knowledge Graph Entities:
    |\color{ForestGreen}Q392: Bob Dylan - American singer-songwriter|
    |\color{ForestGreen}Q62519478: Beatrice Stone|
Knowledge Graph Edges:
    |\color{ForestGreen}Bob Dylan:|
        |\color{ForestGreen}mother:|
            |\color{ForestGreen}Beatrice Stone|

Previous Actions: 
|\color{ForestGreen}EXPAND KG: I should search for the mother of Bob Dylan|
|\color{ForestGreen}SELECT ENTITIES: Q392|
|\color{ForestGreen}SELECT RELATION: P25 - has mother|
|\color{ForestGreen}EXPAND KG: I should search for the mother of Beatrice Stone|

Current task
EXPAND_KG takes two parameters. The first is the entity or entity group to get more information about. Select which entity or entities from the KG to expand.
Provide the QIDs. Options include |\color{BrickRed}[Q392, Q62519478]|
SELECT ENTITIES:
\end{lstlisting}
\end{minipage}

\subsection{Prompt for \textit{selecting-relation} action state}
\label{sec:prompt_sel_rel}

\begin{minipage}{\linewidth}
        \begin{lstlisting}
Original Query: 
    |\color{CornflowerBlue}Who is Bob Dylan's maternal grandmother?|
    
Knowledge Graph Entities:
    |\color{ForestGreen}Q392: Bob Dylan - American singer-songwriter|
    |\color{ForestGreen}Q62519478: Beatrice Stone|
Knowledge Graph Edges:
    |\color{ForestGreen}Bob Dylan:|
        |\color{ForestGreen}mother:|
            |\color{ForestGreen}Beatrice Stone|

Previous Actions: 
|\color{ForestGreen}EXPAND KG: I should search for the mother of Bob Dylan|
|\color{ForestGreen}SELECT ENTITIES: Q392|
|\color{ForestGreen}SELECT RELATION: P25 - has mother|
|\color{ForestGreen}EXPAND KG: I should search for the mother of Beatrice Stone|
|\color{ForestGreen}SELECT ENTITIES: Q62519478|

Your current task is to select the property (PID) to expand along for the selected entities.
The selected entities are: |\color{BrickRed}[Q62519478]|
The options of properties to choose from are:
|\color{BrickRed}P31 - has instance of|
|\color{BrickRed}P21 - has sex of gender|
|\color{BrickRed}P27 - has country of citizenship|
|\color{BrickRed}P735 - has given name|
|\color{BrickRed}...|

Select exactly one property (PID e.g., P10) from those listed above
SELECT PROPERTY:
\end{lstlisting}
\end{minipage}

\subsection{General Prompt for \texttt{evaluate}}
\label{sec:prompt_value}

\begin{minipage}{\linewidth}
        \begin{lstlisting}
Original Query: 
    |\color{CornflowerBlue}Who is Bob Dylan's maternal grandmother?|
    
Knowledge Graph Entities:
    |\color{ForestGreen}Q392: Bob Dylan - American singer-songwriter|
    |\color{ForestGreen}Q62519478: Beatrice Stone|
Knowledge Graph Edges:
    |\color{ForestGreen}Bob Dylan:|
        |\color{ForestGreen}mother:|
            |\color{ForestGreen}Beatrice Stone|

Previous Actions: 
|\color{ForestGreen}EXPAND KG: I should search for the mother of Bob Dylan|
|\color{ForestGreen}SELECT ENTITIES: Q392|
|\color{ForestGreen}SELECT RELATION: P25 - has mother|
|\color{ForestGreen}EXPAND KG: I should search for the mother of Beatrice Stone|
|\color{ForestGreen}SELECT ENTITIES: Q62519478|

Your current task is to evaluate the above knowledge graph and action history.
Based on the original query, the current knowledge graph, and the action history, give the likelihood that the model will correctly answer the question. 
If the most recent action provided information towards the goal and followed the preceding thought, give a high score. 
If the last action was unhelpful, give a low score. 

The output should be a number between 0 and 1 with one decimal. Do not output anything else. 

RATING [0.0-1.0]:
\end{lstlisting}
\end{minipage}

\subsection{Prompt for \texttt{evaluate} on answer (\textit{done})}
\label{sec:prompt_value_done}

\begin{minipage}{\linewidth}
        \begin{lstlisting}
Original Query: 
    |\color{CornflowerBlue}Who is Bob Dylan's maternal grandmother?|
    
Knowledge Graph Entities:
    |\color{ForestGreen}Q392: Bob Dylan - American singer-songwriter|
    |\color{ForestGreen}Q62519478: Beatrice Stone|
    |\color{ForestGreen}Q62519478: Florence Sara Stone|
Knowledge Graph Edges:
    |\color{ForestGreen}Bob Dylan:|
        |\color{ForestGreen}mother:|
            |\color{ForestGreen}Beatrice Stone|
    |\color{ForestGreen}Beatrice Stone:|
        |\color{ForestGreen}mother:|
            |\color{ForestGreen}Florence Sara Stone|

Provided answer: |\color{BrickRed}The answer is Florence Sara Stone.|

Your task is to score the correctness of the provided answer based on the original query, and the knowledge graph.
Give a pessimistic score from 0.0 to 1.0 on how likely the answer is to be correct. 
0.0 if definitely wrong
0.0 if unable to answer based on the knowledge graph
0.5 if unsure
0.7 for probably correct but not confirmed in knowledge graph
1.0 for definitely correct and confirmed in knowledge graph.

Give reasoning to get to the correct answer. Then provide a score.
e.g., 
Reasoning...
So the score for the provided answer should be...
\end{lstlisting}
\end{minipage}

\end{document}